\documentclass{article}

\usepackage{times}
\usepackage{graphicx} 
\usepackage{subfigure}
\usepackage{amsmath}
\usepackage{amssymb}
\usepackage{txfonts}
\usepackage{verbatim}

\usepackage{natbib}

\usepackage{algorithm}
\usepackage{algorithmic}

\usepackage{hyperref}

\usepackage{xcolor}
\definecolor{darkgreen}{rgb}{0,0.6,0}

\usepackage[accepted]{icml2017}

\icmltitlerunning{Parallel Multiscale Autoregressive Density Estimation}

\begin{document} 

\twocolumn[
\icmltitle{Parallel Multiscale Autoregressive Density Estimation}


\begin{icmlauthorlist}
\icmlauthor{Scott Reed}{deepmind}
\icmlauthor{A\"aron van den Oord}{deepmind}
\icmlauthor{Nal Kalchbrenner}{deepmind}
\icmlauthor{Sergio G\'omez Colmenarejo}{deepmind}
\icmlauthor{Ziyu Wang}{deepmind} \\
\icmlauthor{Dan Belov}{deepmind}
\icmlauthor{Nando de Freitas}{deepmind}
\end{icmlauthorlist}


\icmlaffiliation{deepmind}{DeepMind}

\icmlcorrespondingauthor{Scott Reed}{reedscot@google.com}

\icmlkeywords{density models, deep learning}
\vskip 0.3in
]



\printAffiliationsAndNotice{}  

\begin{abstract}
PixelCNN achieves state-of-the-art results in density estimation for natural images.
Although training is fast, inference is costly, requiring one network evaluation per pixel; O(N) for N pixels.
This can be sped up by caching activations, but still involves generating each pixel sequentially.
In this work, we propose a parallelized PixelCNN that allows more efficient inference by modeling certain pixel groups as conditionally independent.
%
%
Our new PixelCNN model achieves competitive density estimation and orders of magnitude speedup - $O(\log N)$ sampling instead of $O(N)$ - enabling the practical generation of $512 \times 512$ images.
We evaluate the model on class-conditional image generation, text-to-image synthesis, and action-conditional video generation, showing that our model achieves the best results among non-pixel-autoregressive density models that allow efficient sampling.
\end{abstract} 

\section{Introduction}
\label{intro}
Many autoregressive image models factorize the joint distribution of images into per-pixel factors:
\begin{align}
\label{eq:factorized}
  p(x_{1:T}) &= \prod_{t=1}^{T}p(x_t | x_{1:t-1})
\end{align}
For example PixelCNN~\citep{oord2016conditional} uses a deep convolutional network with carefully designed filter masking to preserve causal structure, so that all factors in equation~\ref{eq:factorized} can be learned in parallel for a given image.
However, a remaining difficulty is that due to the learned causal structure, inference proceeds sequentially pixel-by-pixel in raster order.

In the naive case, this requires a full network evaluation per pixel.
Caching hidden unit activations can be used to reduce the amount of computation per pixel, as in the 1D case for WaveNet~\cite{oord2016wavenet,ramachandran2017fastgeneration}.
However, even with this optimization, generation is still in serial order by pixel.

Ideally we would generate multiple pixels in parallel, which could greatly accelerate sampling.
In the autoregressive framework this only works if the pixels are modeled as independent.
Thus we need a way to judiciously break weak dependencies among pixels; for example immediately neighboring pixels should not be modeled as independent since they tend to be highly correlated.

%

%

%

\begin{figure}[t!]
\includegraphics[width=\columnwidth]{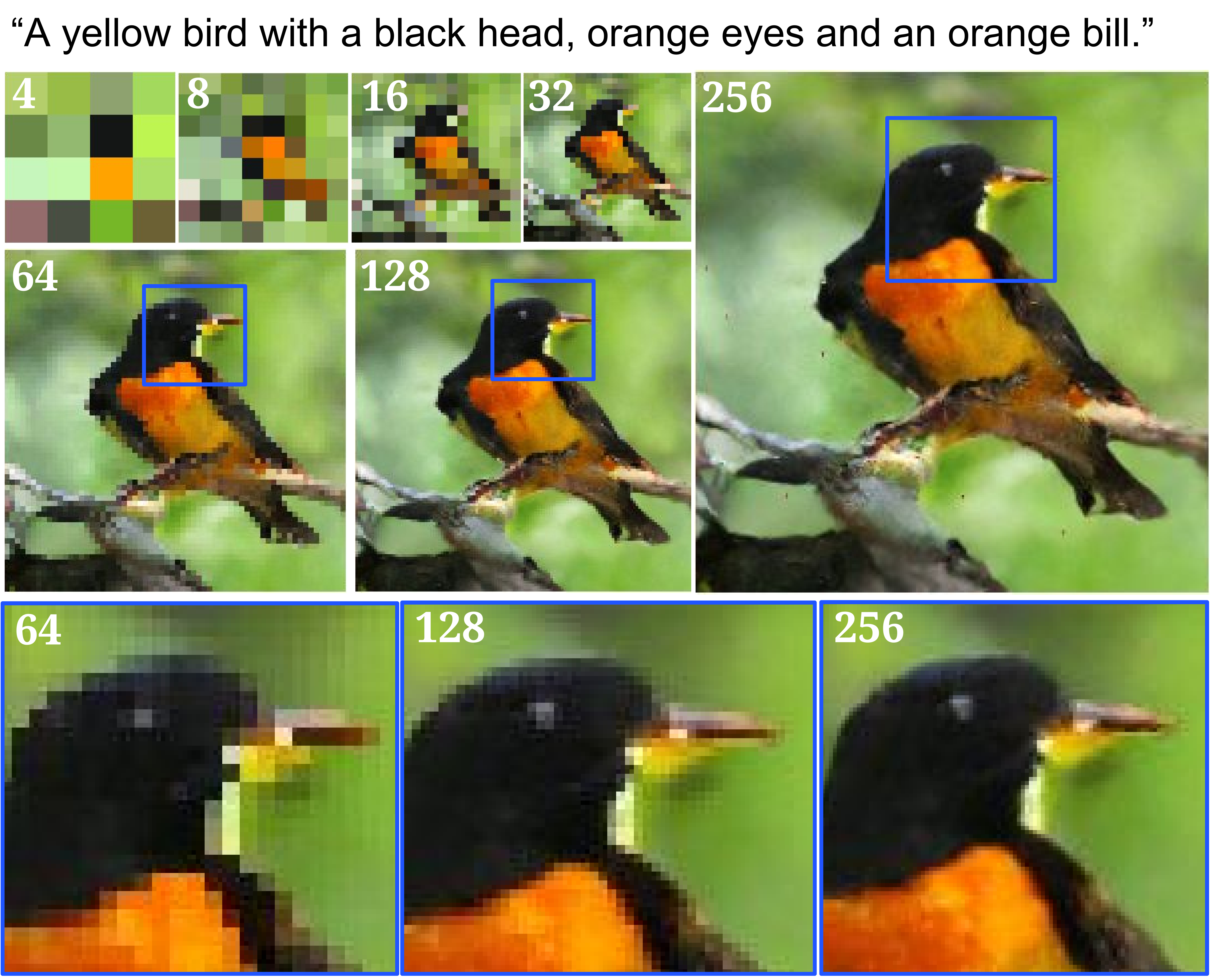}\vspace{-0.15in}
\caption{Samples from our model at resolutions from $4 \times 4$ to $256 \times 256$, conditioned on text and bird part locations in the CUB data set. See Fig.~\ref{fig:cub} and the supplement for more examples.}
\label{fig:concept}
\vspace{-0.1in}
\end{figure}

Multiscale image generation provides one such way to break weak dependencies.
%
In particular, we can model certain groups of pixels as conditionally independent given a lower resolution image and various types of context information, such as preceding frames in a video.
The basic idea is obvious, but nontrivial design problems stand between the idea and a workable implementation.

First, what is the right way to transmit global information from a low-resolution image to each generated pixel of the high-resolution image?
Second, which pixels can we generate in parallel? And given that choice, how can we avoid border artifacts when merging sets of pixels that were generated in parallel, blind to one another?

In this work we show how a very substantial portion of the spatial dependencies in PixelCNN can be cut, with only modest degradation in performance.
Our formulation allows sampling in $O(\log N)$ time for $N$ pixels, instead of $O(N)$ as in the original PixelCNN, resulting in orders of magnitude speedup in practice. 
In the case of video, in which we have access to high-resolution previous frames, we can even sample in $O(1)$ time, with much better performance than comparably-fast baselines.

At a high level, the proposed approach can be viewed as a way to merge per-pixel factors in equation~\ref{eq:factorized}.
If we merge the factors for, e.g. $x_i$ and $x_j$, then that dependency is ``cut'', so the model becomes slightly less expressive.
However, we get the benefit of now being able to sample $x_i$ and $x_j$ in parallel.
%
If we divide the $N$ pixels into $G$ groups of $T$ pixels each, the joint distribution can be written as a product of the corresponding $G$ factors:
%
\begin{align}
\label{eq:factorized2}
p(x^{1:G}_{1:T}) &= \prod_{g=1}^{G}p(x^{(g)}_{1:T} | x^{(1:g-1)}_{1:T})
\end{align}
%
Above we assumed that each of the $G$ groups contains exactly $T$ pixels, but in practice the number can vary.
In this work, we form pixel groups from successively higher-resolution views of an image, arranged into a sub-sampling pyramid, such that $G \in O(\log N)$.

In section~\ref{sec:model} we describe this group structure implemented as a deep convolutional network.
In section~\ref{sec:experiments} we show that the model excels in density estimation and can produce quality high-resolution samples at high speed.
%
%
\section{Related work}
\label{sec:related}
Deep neural autoregressive models have been applied to image generation for many years, showing promise as a tractable yet expressive density model~\citep{larochelle2011neural,uria2013rnade}.
Autoregressive LSTMs have been shown to produce state-of-the-art performance in density estimation on large-scale  datasets such as ImageNet~\cite{Theis2015c,Oord2016pixelRNN}.

Causally-structured convolutional networks such as PixelCNN~\citep{oord2016conditional}
and WaveNet~\citep{oord2016wavenet} improved the speed and scalability of training.
These led to improved autoregressive models for video generation~\citep{kalchbrenner2016video} and machine translation~\citep{kalchbrenner2016neural}.

Non-autoregressive convolutional generator networks have been successful and widely adopted for image generation as well.
Instead of maximizing likelihood, Generative Adversarial Networks (GANs) train a generator network to fool a discriminator network adversary~\cite{GoodfellowPMXWOCB14}.
These networks have been used in a wide variety of conditional image generation schemes such as text and spatial structure to image~\cite{mansimov2015generating,reed2016generative,reed2016learning,wang2016generative}.

The addition of multiscale structure has also been shown to be useful in adversarial networks.
~\citet{denton2015deep} used a Laplacian pyramid to generate images in a coarse-to-fine manner.
~\citet{zhang2016stackgan} composed a low-resolution and high-resolution text-conditional GAN, yielding higher quality $256\times 256$ bird and flower images. 

Generator networks can be combined with a trained model, such as an image classifier or captioning network, to generate high-resolution images via optimization and sampling procedures~\cite{nguyen2016plug}.
%
\citet{Wu2017on} state that it is difficult to quantify GAN performance, and propose Monte Carlo methods to approximate the log-likelihood of GANs on MNIST images.

Both auto-regressive and non auto-regressive deep networks have recently been applied successfully to image super-resolution.
~\citet{shi2016real} developed a sub-pixel convolutional network well-suited to this problem.
\citet{dahl2017pixel} use a PixelCNN as a prior for image super-resolution with a convolutional neural network.
~\citet{johnson2016perceptual} developed a perceptual loss function useful for both style transfer and super-resolution.
GAN variants have also been successful in this domain~\cite{Ledig2016arxiv,sonderby2016amortised}.

Several other deep, tractable density models have recently been developed.
Real NVP~\citep{dinh2016density} learns a mapping from images to a simple noise distribution, which is by construction trivially invertible.
It is built from smaller invertible blocks called coupling layers whose Jacobian is lower-triangular, and also has a multiscale structure.
Inverse Autoregressive Flows~\citep{kingma2016improving} use autoregressive structures in the latent space to learn more flexible posteriors for variational auto-encoders.
Autoregressive models have also been combined with VAEs as decoder models~\citep{gulrajani2016pixelvae}.

The original PixelRNN paper~\citep{Oord2016pixelRNN} actually included a multiscale autoregressive version, in which
PixelRNNs or PixelCNNs were trained at multiple resolutions.
The network producing a given resolution image was conditioned on the image at the next lower resolution.
This work is similarly motivated by the usefulness of multiscale image structure (and the very long history of coarse-to-fine modeling).

Our novel contributions in this work are (1) asymptotically and empirically faster inference by modeling conditional independence structure, (2) scaling to much higher resolution, (3) evaluating the model on a diverse set of challenging benchmarks including class-, text- and structure-conditional image generation and video generation. 

\begin{figure*}[t!]
\centering
\includegraphics[width=0.95\linewidth]{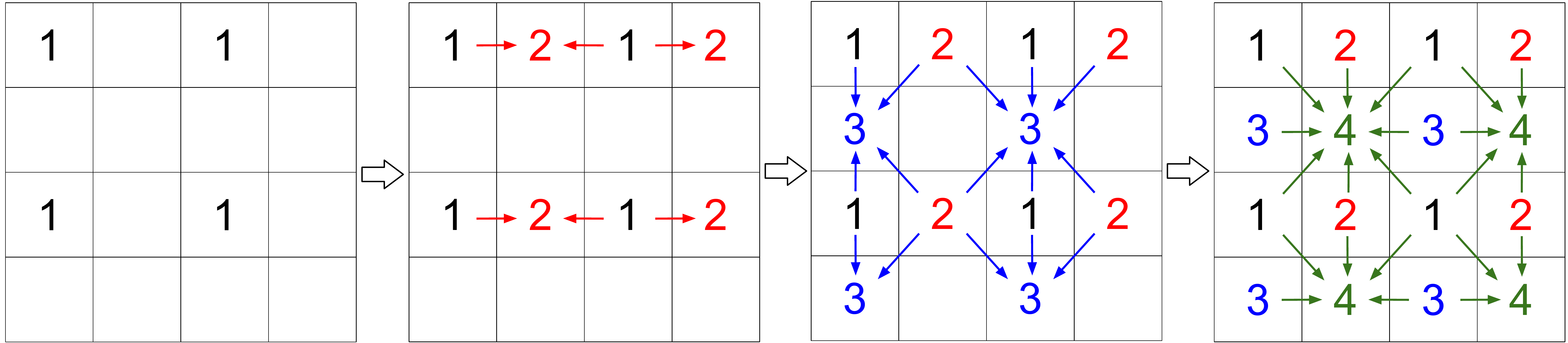}
\vspace{-0.1in}
\caption{Example pixel grouping and ordering for a $4 \times 4$ image. The upper-left corners form group $1$, the upper-right group $2$, and so on. For clarity we only use arrows to indicate immediately-neighboring dependencies, but note that all pixels in preceding groups can be used to predict all pixels in a given group. For example all pixels in group  $2$ can be used to predict pixels in group $4$. In our image experiments pixels in group $1$ originate from a lower-resolution image. For video, they are generated given the previous frames.\label{fig:pixel_groups}}
\vspace{0.15in}
\includegraphics[width=\linewidth]{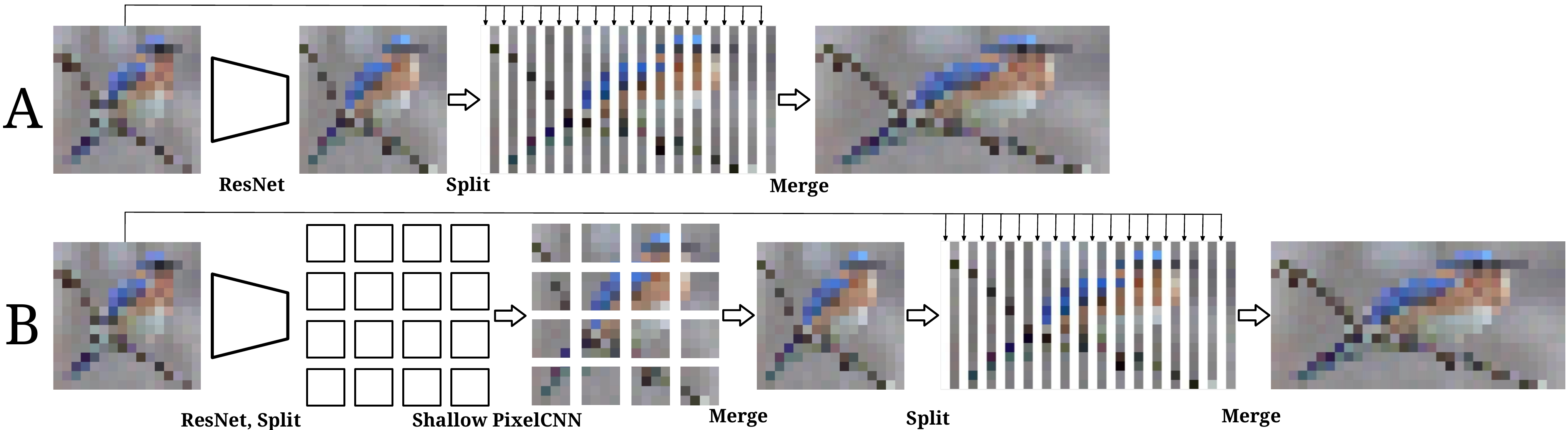}
\vspace{-0.2in}
\caption{A simple form of causal upscaling network, mapping from a $K \times K$ image to $K \times 2K$. The same procedure can be applied in the vertical direction to produce a $2K \times 2K$ image. In reference to figure~\ref{fig:pixel_groups}, the leftmost images could be considered ``group 1'' pixels; i.e. the upper-left corners. The network shown here produces ``group 2'' pixels; i.e. the upper-right corners, completing the top-corners half of the image. \textbf{(A)} In the simplest version, a deep convolutional network (in our case ResNet) directly produces the right image from the left image, and merges column-wise. \textbf{(B)} A more sophisticated version extracts features from a convolutional net, splits the feature map into spatially contiguous blocks, and feeds these in parallel through a shallow PixelCNN. The result is then merged as in (A).\label{fig:upscaling_network}}
\vspace{-0.1in}
\end{figure*}
%
\section{Model}
\label{sec:model}
%
%
%
%
%
%
%

The main design principle that we follow in building the model is a coarse-to-fine ordering of pixels.
Successively higher-resolution frames are generated conditioned on the previous resolution (See for example Figure~\ref{fig:concept}).
Pixels are grouped so as to exploit spatial locality at each resolution, which we describe in detail below.

The training objective is to maximize $\log P(x; \theta)$.
Since the joint distribution factorizes over pixel groups and scales, the training can be trivially parallelized.
%
%
\subsection{Network architecture}
Figure~\ref{fig:pixel_groups} shows how we divide an image into disjoint groups of pixels, with autoregressive structure among the groups.
The key property to notice is that no two adjacent pixels of the high-resolution image are in the same group.
%
%
Also, pixels can depend on other pixels below and to the right, which would have been inaccessible in the standard PixelCNN.
Each group of pixels corresponds to a factor in the joint distribution of equation~\ref{eq:factorized2}.
%

%

%
%
%
%
%
%
%
%
%

Concretely, to create groups we tile the image with $2 \times 2$ blocks.
%
%
The corners of these $2 \times 2$ blocks form the four pixel groups at a given scale; i.e. upper-left, upper-right, lower-left, lower-right.
Note that some pairs of pixels both within each block and also across blocks can still be dependent.
These additional dependencies are important for capturing local textures and avoiding border artifacts.

Figure~\ref{fig:upscaling_network} shows an instantiation of one of these factors as a neural network.
%
%
Similar to the case of PixelCNN, at training time losses and gradients for all of the pixels within a group can be computed in parallel.
At test time, inference proceeds sequentially over pixel groups, in parallel within each group.
Also as in PixelCNN, we model the color channel dependencies - i.e. green sees red, blue sees red and green - using channel masking.

In the case of type-A upscaling networks (See Figure~\ref{fig:upscaling_network}A), sampling each pixel group thus requires $3$ network evaluations~\footnote{However, one could also use a discretized mixture of logistics as output instead of a softmax as in~\citet{salimans2017pixelcnn}, in which case only one network evaluation is needed.}.
In the case of type-B upscaling, the spatial feature map for predicting a group of pixels is divided into contiguous $M \times M$ patches for input to a shallow PixelCNN (See figure~\ref{fig:upscaling_network}B).
This entails $M^2$ very small network evaluations, for each color channel.
We used $M=4$, and the shallow PixelCNN weights are shared across patches.

The division into non-overlapping patches may appear to risk border artifacts when merging.
However, this does not occur for several reasons.
First, each predicted pixel is directly adjacent to several context pixels fed into the upscaling network.
Second, the generated patches are not directly adjacent in the $2K \times 2K$ output image; there is always a row or column of pixels on the border of any pair.

Note that the only learnable portions of the upscaling module are (1) the ResNet encoder of context pixels, and (2) the shallow PixelCNN weights in the case of type-B upscaling.
The ``merge'' and ``split'' operations shown in figure~\ref{fig:upscaling_network} only marshal data and are not associated with parameters.

Given the first group of pixels, the rest of the groups at a given scale can be generated autoregressively.
The first group of pixels can be modeled using the same approach as detailed above, recursively, down to a base resolution at which we use a standard PixelCNN.
At each scale, the number of evaluations is $O(1)$, and the resolution doubles after each upscaling, so the overall complexity is $O(\log N)$ to produce images with $N$ pixels.
%

\begin{figure*}[t!]
\centering
\includegraphics[width=\linewidth]{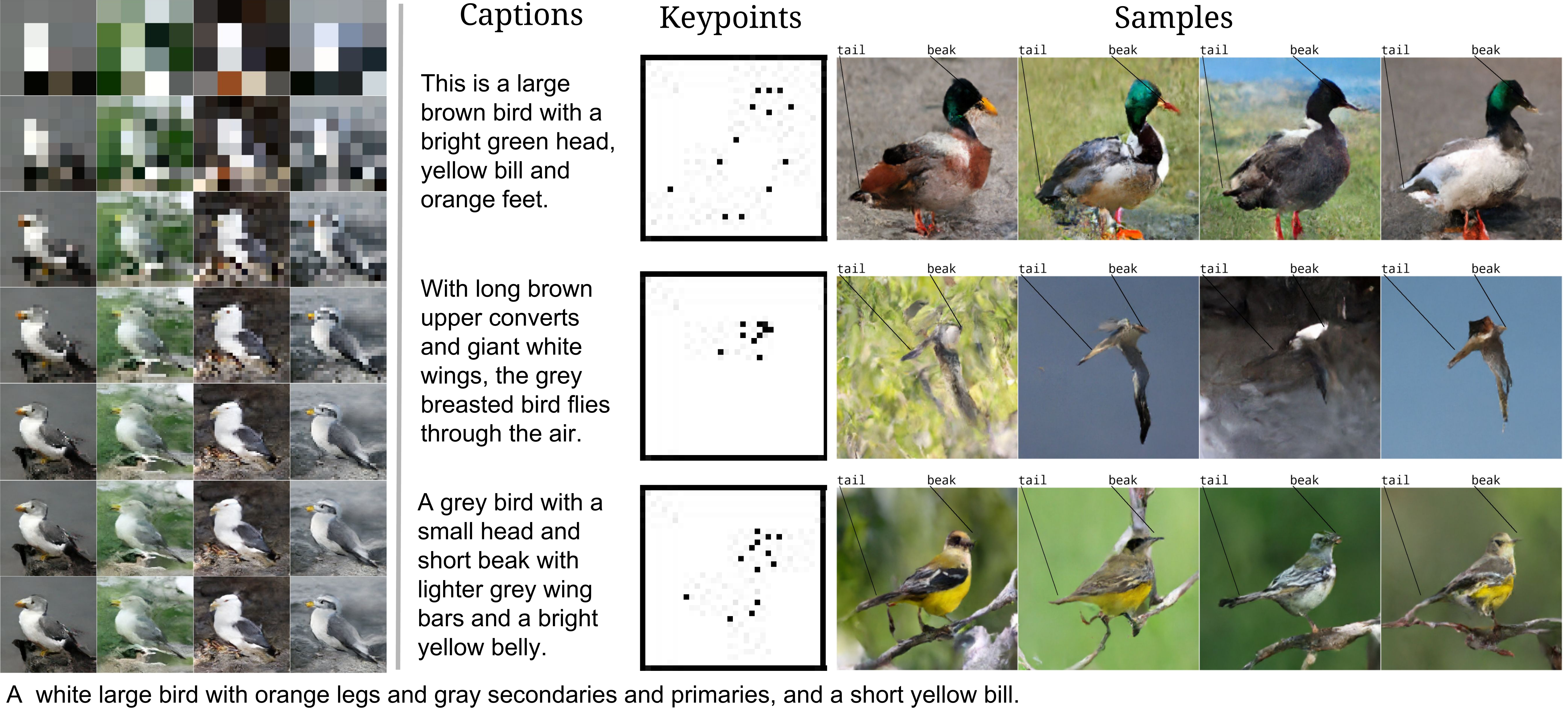}
\vspace{-0.3in}
\caption{Text-to-image bird synthesis. The leftmost column shows the entire sampling process starting by generating $4 \times 4$ images, followed by six upscaling steps, to produce a $256 \times 256$ image. 
The right column shows the final sampled images for several other queries. For each query the associated part keypoints and caption are shown to the left of the samples.\label{fig:cub}}
\vspace{0.1in}
\includegraphics[width=\linewidth]{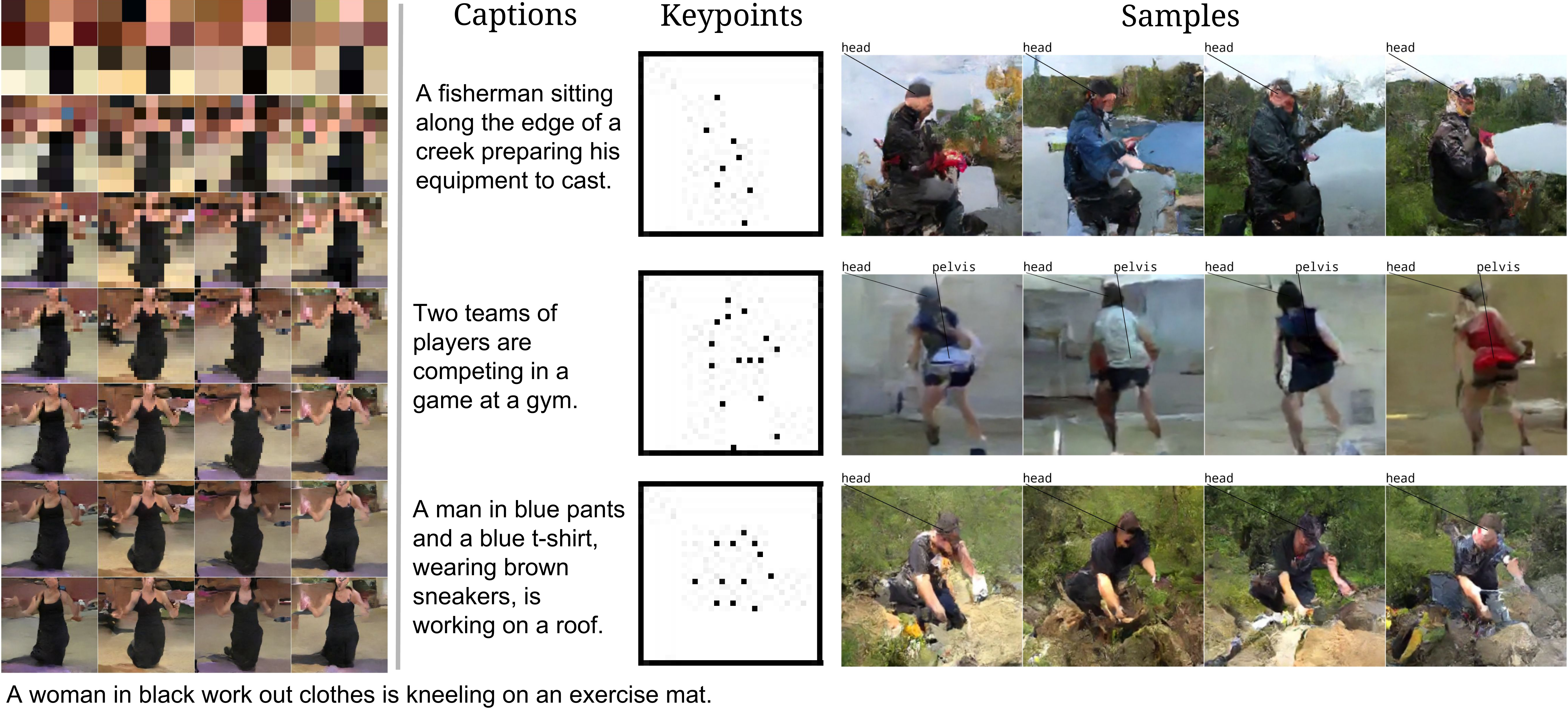}
\vspace{-0.3in}
\caption{Text-to-image human synthesis.The leftmost column again shows the sampling process, and the right column shows the final frame for several more examples. We find that the samples are diverse and usually match the color and position constraints.\label{fig:mpii}}
\vspace{-0.1in}
\end{figure*}

\subsection{Conditional image modeling}
%
%
Given some context information $c$, such as a text description, a segmentation, or previous video frames, we maximize the conditional likelihood $\log P(x | c; \theta)$.
Each factor in equation~\ref{eq:factorized2} simply adds $c$ as an additional conditioning variable.
The upscaling neural network corresponding to each factor takes $c$ as an additional input.

%
For encoding text we used a character-CNN-GRU as in ~\citep{reed2016learning}.
For spatially structured data such as segmentation masks we used a standard convolutional network.
For encoding previous frames in a video we used a ConvLSTM as in~\citep{kalchbrenner2016video}.

\section{Experiments}
\label{sec:experiments}
\subsection{Datasets}
We evaluate our model on ImageNet, Caltech-UCSD Birds (CUB), the MPII Human Pose dataset (MPII), the Microsoft Common Objects in Context dataset (MS-COCO), and the Google Robot Pushing dataset.
\begin{itemize}
\item For ImageNet~\citep{deng2009imagenet}, we trained a class-conditional model using the 1000 leaf node classes.
\item CUB~\citep{wah2011caltech} contains $11,788$ images across $200$ bird species, with $10$ captions
per image. As conditioning information we used a $32 \times 32$ spatial encoding of the 15 annotated bird part locations.
\item MPII~\citep{andriluka20142d} has around $25K$ images of $410$
human activities, with $3$ captions per image.
We kept only the images depicting a single person, and cropped the image centered
around the person, leaving us about $14K$ images. We used a $32 \times 32$ encoding of the 17 annotated human part locations.
\item MS-COCO~\cite{lin2014microsoft} has $80K$ training images with $5$ captions
per image. As conditioning we used the $80$-class segmentation scaled to $32 \times 32$.
\item Robot Pushing~\cite{finn2016unsupervised}  contains sequences of $20$ frames of size $64 \times 64$ showing a robotic arm pushing objects in a basket. There are $50,000$ training sequences and a validation set with the same objects but different arm trajectories. One test set involves a subset of the objects seen during training and another involving novel objects, both captured on an arm and camera viewpoint not seen during training.
\end{itemize}
%
%
All models for ImageNet, CUB, MPII and MS-COCO were trained using RMSprop with hyperparameter 
$\epsilon = 1e-8$, with batch size $128$ for $200K$ steps.
The learning rate was set initially to $1e-4$ and decayed to $1e-5$.

For all of the samples we show, the queries are drawn from the validation split of the corresponding data set.
That is, the captions, key points, segmentation masks, and low-resolution images for super-resolution have not been seen by the model during training.

When we evaluate negative log-likelihood, we only quantize pixel values to $[0,...,255]$ at the target resolution, not separately at each scale.
The lower resolution images are then created by sub-sampling this quantized image.

\subsection{Text and location-conditional generation}

\begin{figure*}[h!]
\centering
\includegraphics[width=\linewidth]{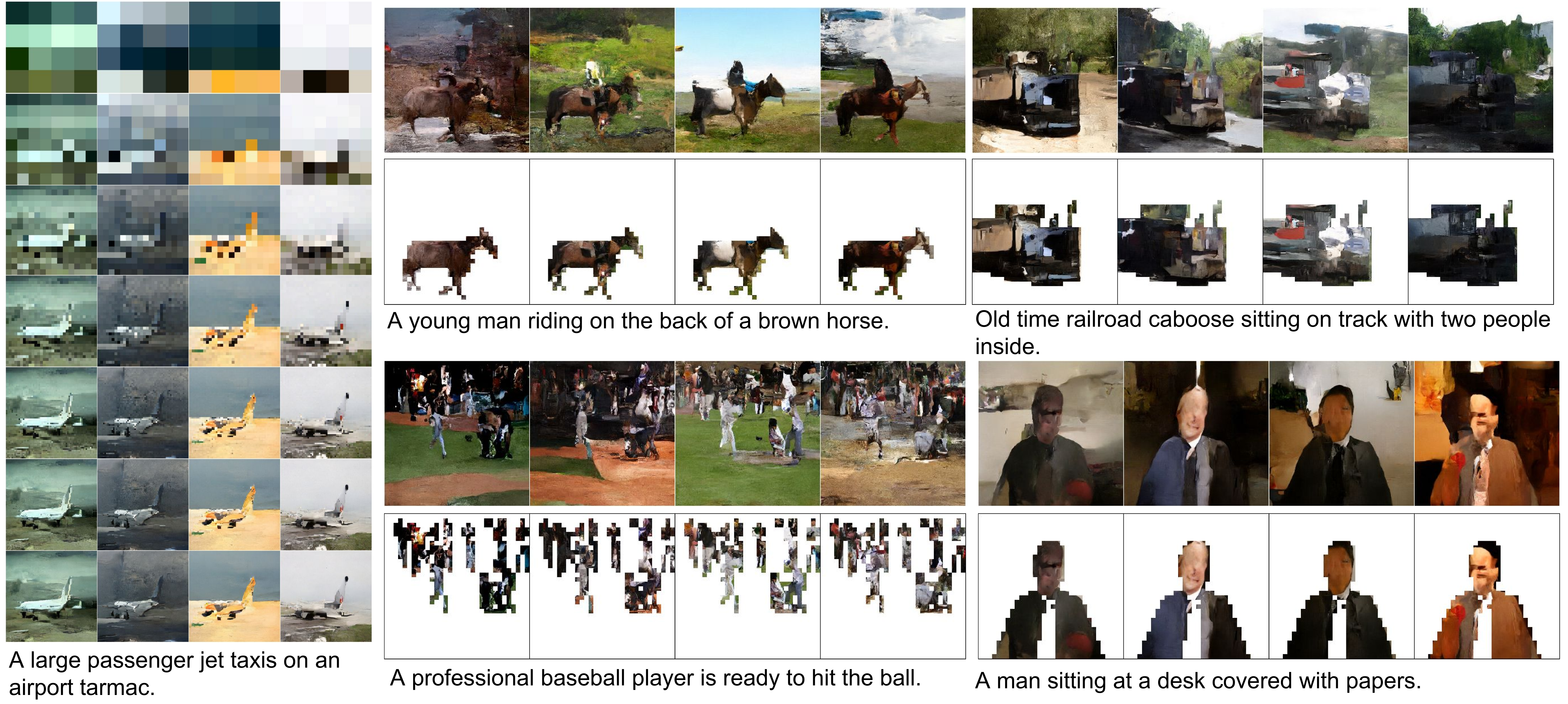}
\vspace{-0.25in}
\caption{Text and segmentation-to-image synthesis. The left column shows the full sampling trajectory from $4 \times 4$ to $256 \times 256$. The caption queries are shown beneath the samples. Beneath each image we show the image masked with the largest object in each scene; i.e. only the foreground pixels in the sample are shown. More samples with all categories masked are included in the supplement.\label{fig:coco}}
\end{figure*}

In this section we show results for CUB, MPII and MS-COCO.
For each dataset we trained type-B upscaling networks with 12 ResNet layers and 4 PixelCNN layers, with 128 hidden units per layer.
The base resolution at which we train a standard PixelCNN was set to $4 \times 4$.

To encode the captions we padded to $201$ characters, then fed into a character-level CNN with three convolutional layers, followed by a GRU and average pooling over time.
Upscaling networks to $8 \times 8$, $16 \times 16$ and $32 \times 32$ shared a single text encoder.
For higher-resolution upscaling networks we trained separate text encoders.
In principle all upscalers could share an encoder, but we trained separably to save memory and time.

For CUB and MPII, we have body part keypoints for birds and humans, respectively.
We encode these into a $32 \times 32 \times P$ binary feature map, where $P$ is the number of parts; $17$ for MPII and $15$ for CUB.
A $1$ indicates the part is visible, and $0$ indicates the part is not visible.
For MS-COCO, we resize the class segmentation mask to $32 \times 32 \times 80$.

For all datasets, we then encode these spatial features using a $12$-layer ResNet.
These features are then depth-concatenated with the text encoding and resized with bilinear interpolation to the spatial size of the image.
If the target resolution for an upscaler network is higher than $32 \times 32$, these conditioning features are randomly cropped along with the target image to a $32 \times 32$ patch.
Because the network is fully convolutional, the network can still generate the full resolution at test time, but we can massively save on memory and computation during training.

Figure~\ref{fig:cub} shows examples of text- and keypoint-to-bird image synthesis.
Figure~\ref{fig:mpii} shows examples of text- and keypoint-to-human image synthesis.
Figure~\ref{fig:coco} shows examples of text- and segmentation-to-image synthesis.


\begin{table}[h!]
\begin{center}
\begin{tabular}{| l | c | c | c |}
\hline
\textbf{CUB} & \textbf{Train} & \textbf{Val} & \textbf{Test} \\
\hline
\hline
PixelCNN & 2.91  & 2.93 & 2.92 \\ \hline
Multiscale PixelCNN & 2.98 & 2.99 & 2.98 \\ 
\hline
\hline
\textbf{MPII} & \textbf{Train} & \textbf{Val} & \textbf{Test} \\ \hline
PixelCNN & 2.90 & 2.92 & 2.92 \\ \hline
Multiscale PixelCNN & 2.91 & 3.03 & 3.03 \\ 
\hline
\hline
\textbf{MS-COCO} & \textbf{Train} & \textbf{Val} & \textbf{Test} \\ \hline
PixelCNN & 3.07 & 3.08 & - \\ \hline
Multiscale PixelCNN & 3.14 & 3.16 & - \\ 
\hline
\end{tabular}
\vspace{-0.1in}
\end{center}
\vspace{-0.05in}
\caption{Text and structure-to image negative conditional log-likelihood in nats per sub-pixel.\label{tab:txt2img}}
\vspace{-0.3in}
\end{table}
Quantitatively, the Multiscale PixelCNN results are not far from those obtained using the original PixelCNN~\citep{reed2016generating}, as shown in Table~\ref{tab:txt2img}.
In addition, we increased the sample resolution by $8\times$.
Qualitatively, the sample quality appears to be on par, but with much greater realism due to the higher resolution.

\begin{figure*}[t!]
\centering
\includegraphics[width=\linewidth]{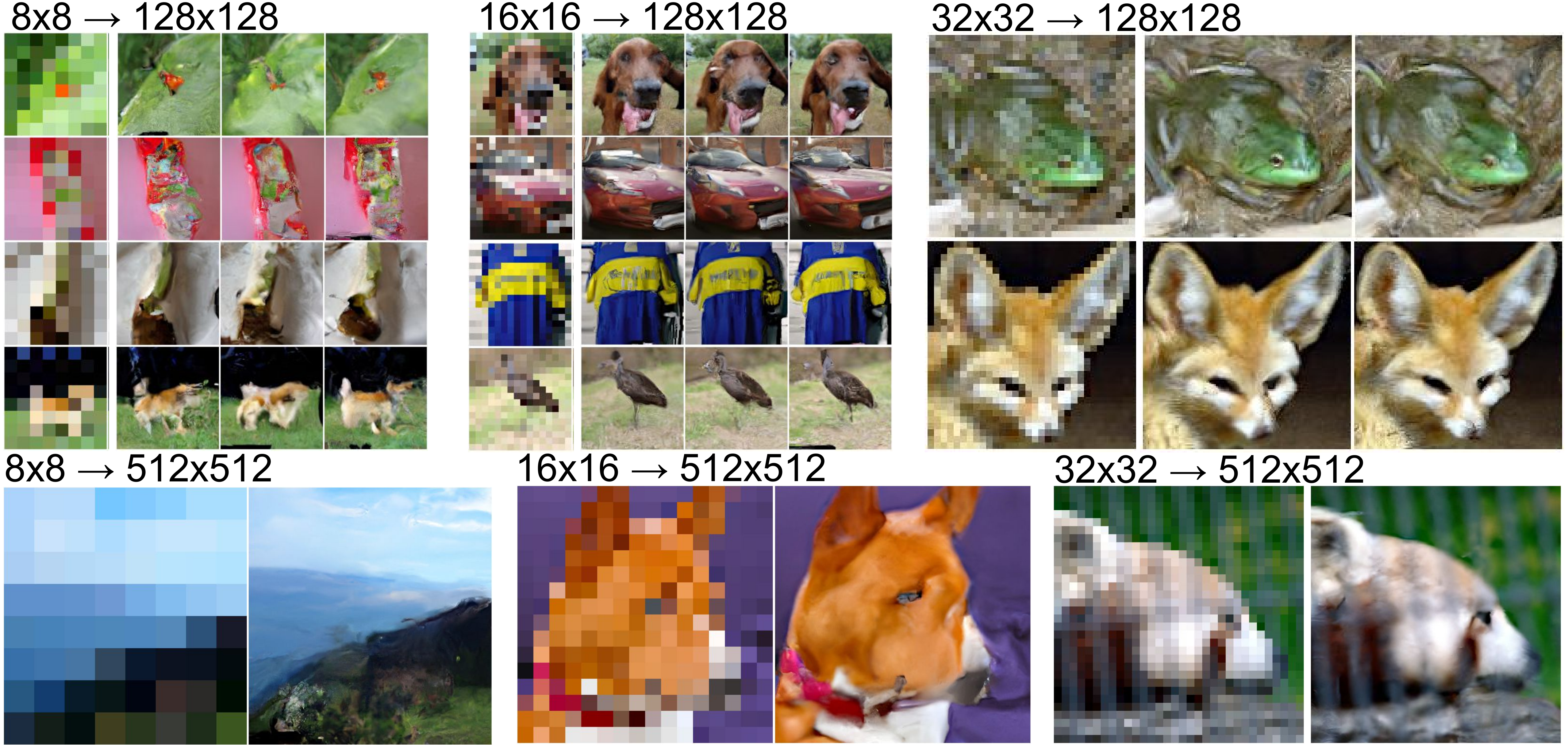}
\vspace{-0.3in}
\caption{Upscaling low-resolution images to $128 \times 128$ and $512 \times 512$. In each group of images, the left column is made of real images, and the right columns of samples from the model.\label{fig:imagenet1}}
\vspace{0.1in}
\includegraphics[width=\linewidth]{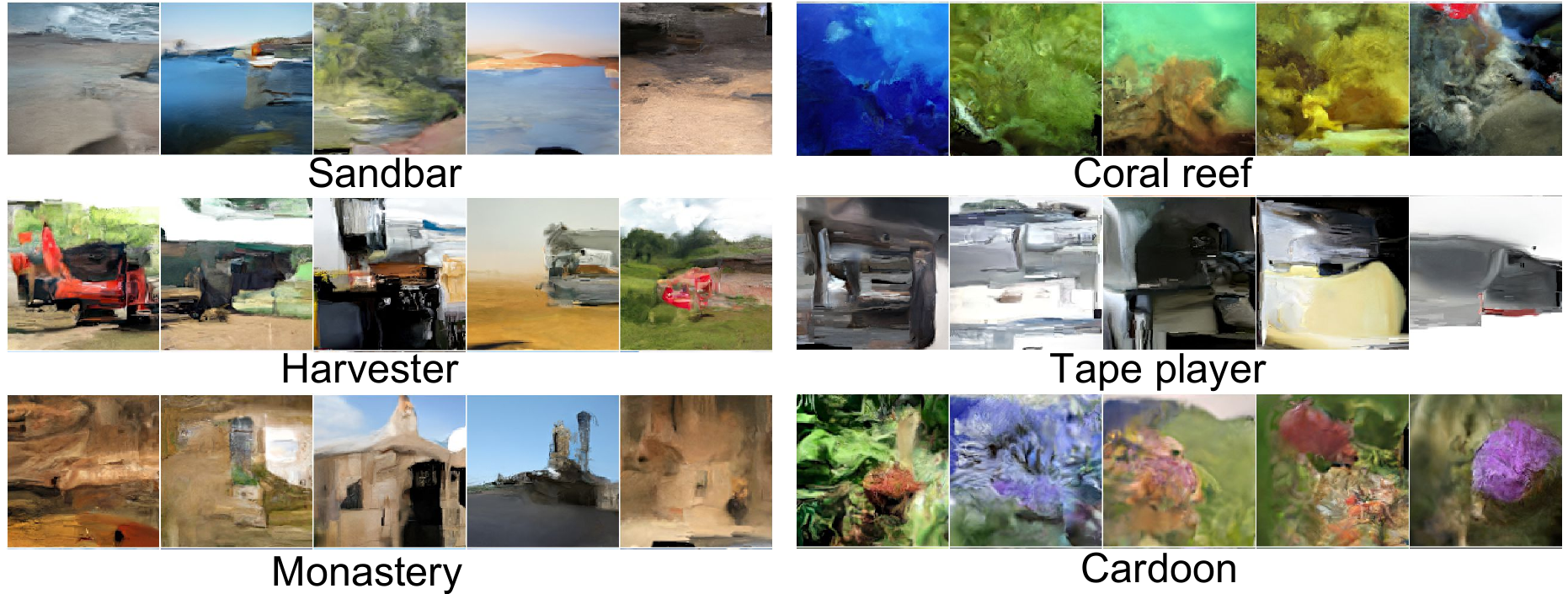}
\vspace{-0.3in}
\caption{Class-conditional $128 \times 128$ samples from a model trained on ImageNet.\label{fig:imagenet2}}
\vspace{-0.1in}
\end{figure*}

\subsection{Action-conditional video generation}
In this section we present results on Robot Pushing videos.
All models were trained to perform future frame prediction conditioned on $2$ starting frames and also on the robot arm actions and state, which are each $5$-dimensional vectors.

We trained two versions of the model, both versions using type-A upscaling networks (See Fig.~\ref{fig:upscaling_network}).
The first is designed to sample in $O(T)$ time, for $T$ video frames.
That is, the number of network evaluations per frame is constant with respect to the number of pixels.

The motivation for training the $O(T)$ model is that previous frames in a video provide very detailed cues for predicting the next frame, so that our pixel groups could be conditionally independent even without access to a low-resolution image.
Without the need to upscale from a low-resolution image, we can produce ``group 1'' pixels - i.e. the upper-left corner group - directly by conditioning on previous frames.
Then a constant number of network evaluations are needed to sample the next three pixel groups at the final scale.

The second version is our multi-step upscaler used in previous experiments, conditioned on both previous frames and robot arm state and actions.
The complexity of sampling from this model is $O(T \log N)$, because at every time step the upscaling procedure must be run, taking $O(\log N)$ time.

The models were trained for $200K$ steps with batch size $64$, using the RMSprop optimizer with centering and $\epsilon = 1e-8$.
The learning rate was initialized to $1e-4$ and decayed by factor $0.3$ after $83K$ steps and after $113K$ steps.
For the $O(T)$ model we used a mixture of discretized logistic outputs~\cite{salimans2017pixelcnn} and for the $O(T\log N)$ model we used a softmax ouptut.
%

%

%
Table~\ref{tab:robot} compares two variants of our model with the original VPN.
Compared to the $O(T)$ baseline - a convolutional LSTM model without spatial dependencies - our $O(T)$ model performs dramatically better.
On the validation set, in which the model needs to generalize to novel combinations of objects and arm trajectories, the $O(T \log N)$ model does much better than our $O(T)$ model, although not as well as the original $O(TN)$ model.

On the testing sets, we observed that the $O(T)$ model performed as well as on the validation set, but the $O(T\log N)$ model showed a drop in performance.
However, this drop does not occur due to the presence of novel objects (in fact this setting actually yields better results), but due to the novel arm and camera configuration used during testing~\footnote{From communication with the Robot Pushing dataset author.}.
It appears that the $O(T \log N)$ model may have overfit to the background details and camera position of the $10$ training arms, but not necessarily to the actual arm and object motions.
It should be possible to overcome this effect with better regularization and perhaps data augmentation such as mirroring and jittering frames, or simply training on data with more diverse camera positions.

The supplement contains example videos generated on the validation set arm trajectories from our $O(T \log N)$ model.
We also trained $64 \rightarrow 128$ and $128 \rightarrow 256$ upscalers conditioned on low-resolution and a previous high-resolution frame, so that we can produce $256 \times 256$ videos.

\subsection{Class-conditional generation}
To compare against other image density models, we trained our Multiscale PixelCNN on ImageNet.
We used type-B upscaling networks (Seee figure~\ref{fig:upscaling_network}) with 12 ResNet~\cite{he2016identity} layers and 4 PixelCNN layers, with 256 hidden units per layer.
For all PixelCNNs in the model, we used the same architecture as in~\citep{oord2016conditional}.
We generated images with a base resolution of $8 \times 8$ and trained four upscaling networks to produce up to $128 \times 128$ samples.
At scales $64 \times 64$ and above, during training we randomly cropped the image to $32 \times 32$.
This accelerates training but does not pose a problem at test time because all of the networks are fully convolutional.

\begin{table}[t!]
\begin{center}
\begin{tabular}{| l | c | c | c | c |}
\hline
\textbf{Model} & \textbf{Tr} & \textbf{Val} & \textbf{Ts-seen} & \textbf{Ts-novel} \\
\hline
\hline
O(T) baseline & - & 2.06 & 2.08 & 2.07 \\ \hline
O(TN) VPN & - & 0.62 & 0.64 & 0.64 \\ 
\hline
\hline
O(T) VPN & 1.03 & 1.04 & 1.04 & 1.04 \\ \hline
O(T log N) VPN & 0.74 & 0.74 & 1.06 & 0.97 \\ \hline
\end{tabular}
\end{center}
\vspace{-0.1in}
\caption{Robot videos neg. log-likelihood in nats per sub-pixel. ``Tr'' is the training set, ``Ts-seen'' is the test set with novel arm and camera configuration and previously seen objects, and ``Ts-novel'' is the same as ``Ts-seen'' but with novel objects.\label{tab:robot}}
\vspace{-0.1in}
\end{table}

Table~\ref{tab:imagenet} shows the results.
%
%
On both $32 \times 32$ and $64 \times 64$ ImageNet it achieves significantly better likelihood scores than have been reported for any non-pixel-autoregressive density models, such as ConvDRAW and Real NVP, that also allow efficient sampling.

Of course, performance of these approaches varies considerably depending on the implementation details, especially in the design and capacity of deep neural networks used.
But it is notable that the very simple and direct approach developed here can surpass the state-of-the-art among fast-sampling density models.


\begin{table}[h!]
\begin{center}
\begin{tabular}{| l | c | c | c |}
\hline
\textbf{Model} & \textbf{32} & \textbf{64} & \textbf{128} \\
\hline
\hline
PixelRNN & 3.86 (3.83) & 3.64(3.57) & - \\ \hline
PixelCNN & 3.83 (3.77) & 3.57(3.48) & - \\ \hline
\hline
Real NVP & 4.28(4.26) & 3.98(3.75) & - \\ \hline
Conv. DRAW & 4.40(4.35) & 4.10(4.04) & - \\
\hline
\hline
Ours & 3.95(3.92) & 3.70(3.67) & 3.55(3.42) \\ \hline
\end{tabular}
\end{center}
\vspace{-0.15in}
\caption{ImageNet negative log-likelihood in bits per sub-pixel at $32 \times 32$, $64 \times 64$ and $128 \times 128$ resolution.\label{tab:imagenet}}
\end{table}

In Figure~\ref{fig:imagenet2} we show examples of diverse $128 \times 128$ class conditional image generation. 

Interestingly, the model often produced quite realistic bird images from scratch when trained on CUB, and these samples looked more realistic than any animal image generated by our ImageNet models.
%
One plausible explanation for this difference is a lack of model capacity; a single network modeling the $1000$ very diverse ImageNet categories can devote only very limited capacity to each one, compared to a network that only needs to model birds.
This suggests that finding ways to increase capacity without slowing down training or sampling could be a promising direction.

Figure~\ref{fig:imagenet1} shows upscaling starting from ground-truth images of size $8 \times 8$, $16 \times 16$ and $32 \times 32$.
We observe the largest diversity of samples in terms of global structure when starting from $8 \times 8$, but less realistic results due to the more challenging nature of the problem.
Upscaling starting from $32 \times 32$ results in much more realistic images.
Here the diversity is apparent in the samples (as in the data, conditioned on low-resolution) in the local details such as the dog's fur patterns or the frog's eye contours.

\subsection{Sampling time comparison}
%
\begin{table}
\begin{center}
\begin{tabular}{| l | c |  c | c |}
\hline
\textbf{Model} & \textbf{scale} & \textbf{time} & \textbf{speedup} \\
\hline
\hline
\small{$O(N)$ PixelCNN} & $32$ & $120.0$ & $1.0 \times$ \\
\hline
\small{$O(\log N)$ PixelCNN} & $32$ & $1.17$ & $102 \times$ \\
\hline
\small{$O(\log N)$ PixelCNN}, in-graph & $32$ & $1.14$ & $105 \times$ \\
\hline
\hline
\small{$O(TN)$ VPN} & $64$ & $1929.8$ & $1.0 \times$\\
\hline
\small{$O(T)$ VPN} & $64$ & $0.38$ & $5078 \times$\\
\hline
\small{$O(T)$ VPN}, in-graph & $64$ & $0.37$ & $5215 \times$ \\
\hline
\small{$O(T \log N)$ VPN} & $64$ & $3.82$ & $505 \times$ \\
\hline
\small{$O(T \log N)$ VPN}, in-graph & $64$ & $3.07$ & $628 \times$\\
\hline
\end{tabular}
\end{center}
\vspace{-0.15in}
\caption{Sampling speed of several models in seconds per frame on an Nvidia Quadro M4000 GPU. The top three rows were measured on $32 \times 32$ ImageNet, with batch size of 30. The bottom five rows were measured on generating $64 \times 64$ videos of $18$ frames each, averaged over $5$ videos.\label{tab:speed}}
\vspace{-0.2in}
\end{table}

As expected, we observe a very large speedup of our model compared to sampling from a standard PixelCNN at the same resolution (see Table~\ref{tab:speed}).
Even at $32 \times 32$ we observe two orders of magnitude speedup, and the speedup is greater for higher resolution.
%

%
Since our model only requires $O(\log N)$ network evaluations to sample, we can fit the entire computation graph for sampling into memory, for reasonable batch sizes.
In-graph computation in TensorFlow can further improve the speed of both image and video generation, due to 
 reduced overhead by avoiding repeated calls to \texttt{sess.run}.

Since our model has a PixelCNN at the lowest resolution, it can also be accelerated by caching PixelCNN hidden unit activations, recently implemented b by~\citet{ramachandran2017fastgeneration}.
This could allow one to use higher-resolution base PixelCNNs without sacrificing speed.




\section{Conclusions}
In this paper, we developed a parallelized, multiscale version of PixelCNN.
It achieves competitive density estimation results on
CUB, MPII, MS-COCO, ImageNet, and Robot Pushing videos, surpassing all other density models that admit fast sampling.
Qualitatively, it can achieve compelling results in text-to-image synthesis and video generation, as well as diverse super-resolution from very small images all the way to $512 \times 512$.

Many more samples from all of our models can be found in the appendix and supplementary material.
%

\bibliography{references}

\begin{thebibliography}{34}
\providecommand{\natexlab}[1]{#1}
\providecommand{\url}[1]{\texttt{#1}}
\expandafter\ifx\csname urlstyle\endcsname\relax
  \providecommand{\doi}[1]{doi: #1}\else
  \providecommand{\doi}{doi: \begingroup \urlstyle{rm}\Url}\fi

\bibitem[Andriluka et~al.(2014)Andriluka, Pishchulin, Gehler, and
  Schiele]{andriluka20142d}
Andriluka, Mykhaylo, Pishchulin, Leonid, Gehler, Peter, and Schiele, Bernt.
\newblock 2d human pose estimation: New benchmark and state of the art
  analysis.
\newblock In \emph{CVPR}, pp.\  3686--3693, 2014.

\bibitem[Dahl et~al.(2017)Dahl, Norouzi, and Shlens]{dahl2017pixel}
Dahl, Ryan, Norouzi, Mohammad, and Shlens, Jonathon.
\newblock Pixel recursive super resolution.
\newblock \emph{arXiv preprint arXiv:1702.00783}, 2017.

\bibitem[Deng et~al.(2009)Deng, Dong, Socher, Li, Li, and
  Fei-Fei]{deng2009imagenet}
Deng, Jia, Dong, Wei, Socher, Richard, Li, Li-Jia, Li, Kai, and Fei-Fei, Li.
\newblock {ImageNet}: A large-scale hierarchical image database.
\newblock In \emph{CVPR}, 2009.

\bibitem[Denton et~al.(2015)Denton, Chintala, Szlam, and
  Fergus]{denton2015deep}
Denton, Emily~L, Chintala, Soumith, Szlam, Arthur, and Fergus, Rob.
\newblock Deep generative image models using a {Laplacian} pyramid of
  adversarial networks.
\newblock In \emph{NIPS}, pp.\  1486--1494, 2015.

\bibitem[Dinh et~al.(2016)Dinh, Sohl-Dickstein, and Bengio]{dinh2016density}
Dinh, Laurent, Sohl-Dickstein, Jascha, and Bengio, Samy.
\newblock Density estimation using {Real NVP}.
\newblock In \emph{NIPS}, 2016.

\bibitem[Finn et~al.(2016)Finn, Goodfellow, and Levine]{finn2016unsupervised}
Finn, Chelsea, Goodfellow, Ian, and Levine, Sergey.
\newblock Unsupervised learning for physical interaction through video
  prediction.
\newblock In \emph{NIPS}, 2016.

\bibitem[Goodfellow et~al.(2014)Goodfellow, Pouget{-}Abadie, Mirza, Xu,
  Warde{-}Farley, Ozair, Courville, and Bengio]{GoodfellowPMXWOCB14}
Goodfellow, Ian~J., Pouget{-}Abadie, Jean, Mirza, Mehdi, Xu, Bing,
  Warde{-}Farley, David, Ozair, Sherjil, Courville, Aaron~C., and Bengio,
  Yoshua.
\newblock Generative adversarial nets.
\newblock In \emph{NIPS}, 2014.

\bibitem[Gulrajani et~al.(2016)Gulrajani, Kumar, Ahmed, Taiga, Visin, Vazquez,
  and Courville]{gulrajani2016pixelvae}
Gulrajani, Ishaan, Kumar, Kundan, Ahmed, Faruk, Taiga, Adrien~Ali, Visin,
  Francesco, Vazquez, David, and Courville, Aaron.
\newblock {PixelVAE}: A latent variable model for natural images.
\newblock \emph{arXiv preprint arXiv:1611.05013}, 2016.

\bibitem[He et~al.(2016)He, Zhang, Ren, and Sun]{he2016identity}
He, Kaiming, Zhang, Xiangyu, Ren, Shaoqing, and Sun, Jian.
\newblock Identity mappings in deep residual networks.
\newblock In \emph{ECCV}, pp.\  630--645, 2016.

\bibitem[Johnson et~al.(2016)Johnson, Alahi, and
  Fei-Fei]{johnson2016perceptual}
Johnson, Justin, Alahi, Alexandre, and Fei-Fei, Li.
\newblock Perceptual losses for real-time style transfer and super-resolution.
\newblock In \emph{ECCV}, 2016.

\bibitem[Kalchbrenner et~al.(2016{\natexlab{a}})Kalchbrenner, Espeholt,
  Simonyan, Oord, Graves, and Kavukcuoglu]{kalchbrenner2016neural}
Kalchbrenner, Nal, Espeholt, Lasse, Simonyan, Karen, Oord, Aaron van~den,
  Graves, Alex, and Kavukcuoglu, Koray.
\newblock Neural machine translation in linear time.
\newblock \emph{arXiv preprint arXiv:1610.10099}, 2016{\natexlab{a}}.

\bibitem[Kalchbrenner et~al.(2016{\natexlab{b}})Kalchbrenner, Oord, Simonyan,
  Danihelka, Vinyals, Graves, and Kavukcuoglu]{kalchbrenner2016video}
Kalchbrenner, Nal, Oord, Aaron van~den, Simonyan, Karen, Danihelka, Ivo,
  Vinyals, Oriol, Graves, Alex, and Kavukcuoglu, Koray.
\newblock Video pixel networks.
\newblock \emph{Preprint arXiv:1610.00527}, 2016{\natexlab{b}}.

\bibitem[Kingma \& Salimans(2016)Kingma and Salimans]{kingma2016improving}
Kingma, Diederik~P and Salimans, Tim.
\newblock Improving variational inference with inverse autoregressive flow.
\newblock In \emph{NIPS}, 2016.

\bibitem[Larochelle \& Murray(2011)Larochelle and Murray]{larochelle2011neural}
Larochelle, Hugo and Murray, Iain.
\newblock The neural autoregressive distribution estimator.
\newblock In \emph{AISTATS}, 2011.

\bibitem[Ledig et~al.(2016)Ledig, Theis, Huszar, Caballero, Cunningham, Acosta,
  Aitken, Tejani, Totz, Wang, and Shi]{Ledig2016arxiv}
Ledig, Christian, Theis, Lucas, Huszar, Ferenc, Caballero, Jose, Cunningham,
  Andrew, Acosta, Alejandro, Aitken, Andrew, Tejani, Alykhan, Totz, Johannes,
  Wang, Zehan, and Shi, Wenzhe.
\newblock Photo-realistic single image super-resolution using a generative
  adversarial network.
\newblock 2016.

\bibitem[Lin et~al.(2014)Lin, Maire, Belongie, Hays, Perona, Ramanan,
  Doll{\'a}r, and Zitnick]{lin2014microsoft}
Lin, Tsung-Yi, Maire, Michael, Belongie, Serge, Hays, James, Perona, Pietro,
  Ramanan, Deva, Doll{\'a}r, Piotr, and Zitnick, C~Lawrence.
\newblock Microsoft {COCO: Common} objects in context.
\newblock In \emph{ECCV}, pp.\  740--755, 2014.

\bibitem[Mansimov et~al.(2015)Mansimov, Parisotto, Ba, and
  Salakhutdinov]{mansimov2015generating}
Mansimov, Elman, Parisotto, Emilio, Ba, Jimmy~Lei, and Salakhutdinov, Ruslan.
\newblock Generating images from captions with attention.
\newblock In \emph{ICLR}, 2015.

\bibitem[Nguyen et~al.(2016)Nguyen, Yosinski, Bengio, Dosovitskiy, and
  Clune]{nguyen2016plug}
Nguyen, Anh, Yosinski, Jason, Bengio, Yoshua, Dosovitskiy, Alexey, and Clune,
  Jeff.
\newblock Plug \& play generative networks: Conditional iterative generation of
  images in latent space.
\newblock \emph{arXiv preprint arXiv:1612.00005}, 2016.

\bibitem[Oord et~al.(2016)Oord, Dieleman, Zen, Simonyan, Vinyals, Graves,
  Kalchbrenner, Senior, and Kavukcuoglu]{oord2016wavenet}
Oord, Aaron van~den, Dieleman, Sander, Zen, Heiga, Simonyan, Karen, Vinyals,
  Oriol, Graves, Alex, Kalchbrenner, Nal, Senior, Andrew, and Kavukcuoglu,
  Koray.
\newblock Wavenet: A generative model for raw audio.
\newblock \emph{arXiv preprint arXiv:1609.03499}, 2016.

\bibitem[Ramachandran et~al.(2017)Ramachandran, Paine, Khorrami, Babaeizadeh,
  Chang, Zhang, Hasegawa-Johnson, Campbell, and
  Huang]{ramachandran2017fastgeneration}
Ramachandran, Prajit, Paine, Tom~Le, Khorrami, Pooya, Babaeizadeh, Mohammad,
  Chang, Shiyu, Zhang, Yang, Hasegawa-Johnson, Mark, Campbell, Roy, and Huang,
  Thomas.
\newblock Fast generation for convolutional autoregressive models.
\newblock 2017.

\bibitem[Reed et~al.(2016{\natexlab{a}})Reed, Akata, Mohan, Tenka, Schiele, and
  Lee]{reed2016learning}
Reed, Scott, Akata, Zeynep, Mohan, Santosh, Tenka, Samuel, Schiele, Bernt, and
  Lee, Honglak.
\newblock Learning what and where to draw.
\newblock In \emph{NIPS}, 2016{\natexlab{a}}.

\bibitem[Reed et~al.(2016{\natexlab{b}})Reed, Akata, Yan, Logeswaran, Schiele,
  and Lee]{reed2016generative}
Reed, Scott, Akata, Zeynep, Yan, Xinchen, Logeswaran, Lajanugen, Schiele,
  Bernt, and Lee, Honglak.
\newblock Generative adversarial text-to-image synthesis.
\newblock In \emph{ICML}, 2016{\natexlab{b}}.

\bibitem[Reed et~al.(2016{\natexlab{c}})Reed, van~den Oord, Kalchbrenner,
  Bapst, Botvinick, and de~Freitas]{reed2016generating}
Reed, Scott, van~den Oord, A\"aron, Kalchbrenner, Nal, Bapst, Victor,
  Botvinick, Matt, and de~Freitas, Nando.
\newblock Generating interpretable images with controllable structure.
\newblock Technical report, 2016{\natexlab{c}}.

\bibitem[Salimans et~al.(2017)Salimans, Karpathy, Chen, and
  Kingma]{salimans2017pixelcnn}
Salimans, Tim, Karpathy, Andrej, Chen, Xi, and Kingma, Diederik~P.
\newblock {PixelCNN++}: Improving the {PixelCNN} with discretized logistic
  mixture likelihood and other modifications.
\newblock \emph{arXiv preprint arXiv:1701.05517}, 2017.

\bibitem[Shi et~al.(2016)Shi, Caballero, Husz{\'a}r, Totz, Aitken, Bishop,
  Rueckert, and Wang]{shi2016real}
Shi, Wenzhe, Caballero, Jose, Husz{\'a}r, Ferenc, Totz, Johannes, Aitken,
  Andrew~P, Bishop, Rob, Rueckert, Daniel, and Wang, Zehan.
\newblock Real-time single image and video super-resolution using an efficient
  sub-pixel convolutional neural network.
\newblock In \emph{CVPR}, 2016.

\bibitem[S{\o}nderby et~al.(2017)S{\o}nderby, Caballero, Theis, Shi, and
  Husz{\'a}r]{sonderby2016amortised}
S{\o}nderby, Casper~Kaae, Caballero, Jose, Theis, Lucas, Shi, Wenzhe, and
  Husz{\'a}r, Ferenc.
\newblock Amortised {MAP} inference for image super-resolution.
\newblock 2017.

\bibitem[Theis \& Bethge(2015)Theis and Bethge]{Theis2015c}
Theis, L. and Bethge, M.
\newblock Generative image modeling using spatial {LSTMs}.
\newblock In \emph{NIPS}, 2015.

\bibitem[Uria et~al.(2013)Uria, Murray, and Larochelle]{uria2013rnade}
Uria, Benigno, Murray, Iain, and Larochelle, Hugo.
\newblock {RNADE}: The real-valued neural autoregressive density-estimator.
\newblock In \emph{NIPS}, 2013.

\bibitem[van~den Oord et~al.(2016{\natexlab{a}})van~den Oord, Kalchbrenner, and
  Kavukcuoglu]{Oord2016pixelRNN}
van~den Oord, A{\"{a}}ron, Kalchbrenner, Nal, and Kavukcuoglu, Koray.
\newblock Pixel recurrent neural networks.
\newblock In \emph{ICML}, pp.\  1747--1756, 2016{\natexlab{a}}.

\bibitem[van~den Oord et~al.(2016{\natexlab{b}})van~den Oord, Kalchbrenner,
  Vinyals, Espeholt, Graves, and Kavukcuoglu]{oord2016conditional}
van~den Oord, A{\"{a}}ron, Kalchbrenner, Nal, Vinyals, Oriol, Espeholt, Lasse,
  Graves, Alex, and Kavukcuoglu, Koray.
\newblock Conditional image generation with {PixelCNN} decoders.
\newblock In \emph{NIPS}, 2016{\natexlab{b}}.

\bibitem[Wah et~al.(2011)Wah, Branson, Welinder, Perona, and
  Belongie]{wah2011caltech}
Wah, Catherine, Branson, Steve, Welinder, Peter, Perona, Pietro, and Belongie,
  Serge.
\newblock The {Caltech-UCSD} birds-200-2011 dataset.
\newblock 2011.

\bibitem[Wang \& Gupta(2016)Wang and Gupta]{wang2016generative}
Wang, Xiaolong and Gupta, Abhinav.
\newblock Generative image modeling using style and structure adversarial
  networks.
\newblock In \emph{ECCV}, pp.\  318--335, 2016.

\bibitem[Wu et~al.(2017)Wu, Burda, Salakhutdinov, and Grosse]{Wu2017on}
Wu, Yuhuai, Burda, Yuri, Salakhutdinov, Ruslan, and Grosse, Roger.
\newblock On the quantitative analysis of decoder-based generative models.
\newblock 2017.

\bibitem[Zhang et~al.(2016)Zhang, Xu, Li, Zhang, Huang, Wang, and
  Metaxas]{zhang2016stackgan}
Zhang, Han, Xu, Tao, Li, Hongsheng, Zhang, Shaoting, Huang, Xiaolei, Wang,
  Xiaogang, and Metaxas, Dimitris.
\newblock {StackGAN}: Text to photo-realistic image synthesis with stacked
  generative adversarial networks.
\newblock \emph{arXiv preprint arXiv:1612.03242}, 2016.

\end{thebibliography}
\bibliographystyle{icml2017}

\newpage
\newpage

\section{Appendix}
Below we show additional samples.

\begin{figure*}[h]
\centering
\includegraphics[width=\linewidth]{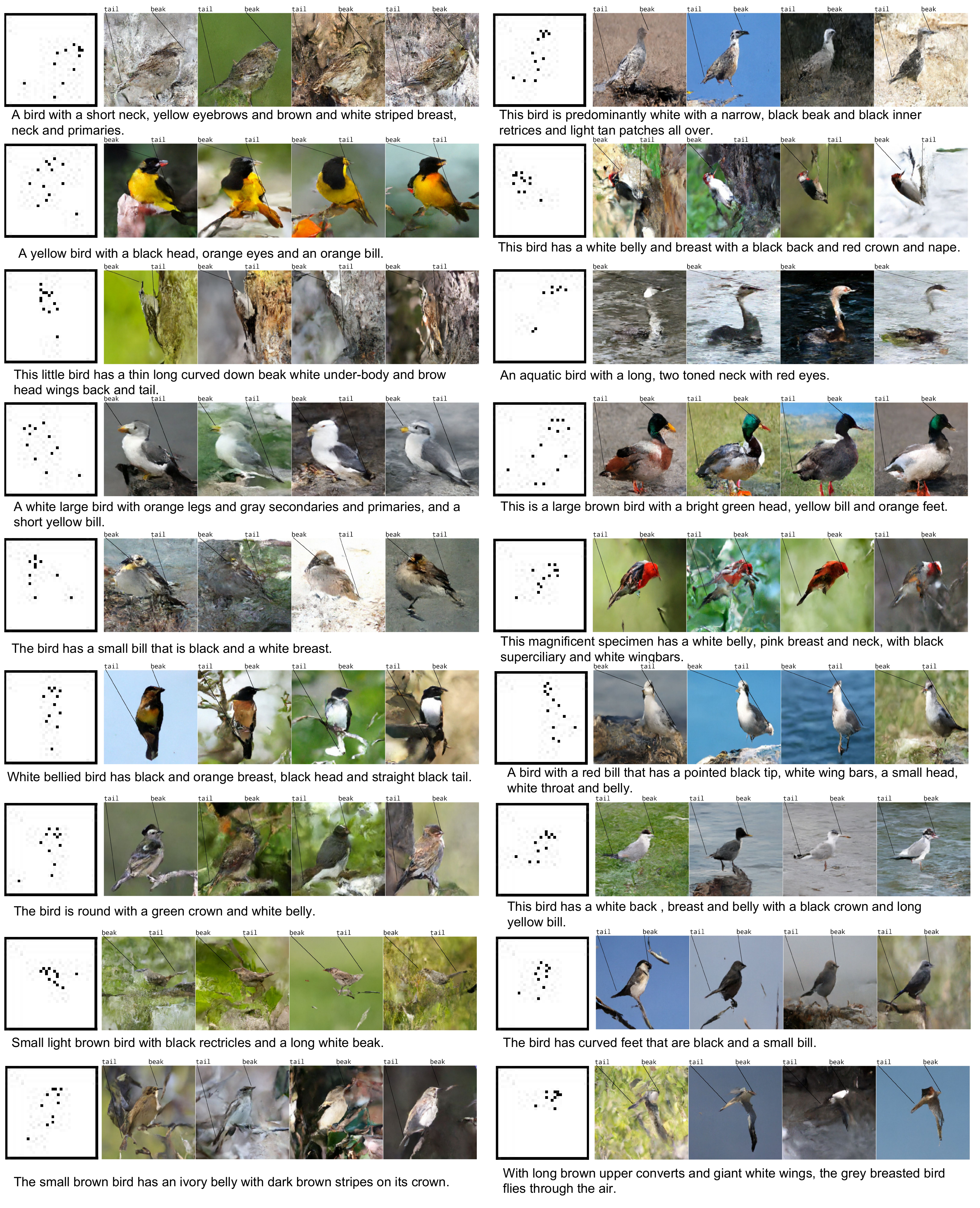}
\vspace{-0.15in}
\caption{Additional CUB samples randomly chosen from the validation set.\label{fig:cub_appendix}}
\end{figure*}

\begin{figure*}[h]
\centering
\includegraphics[width=\linewidth]{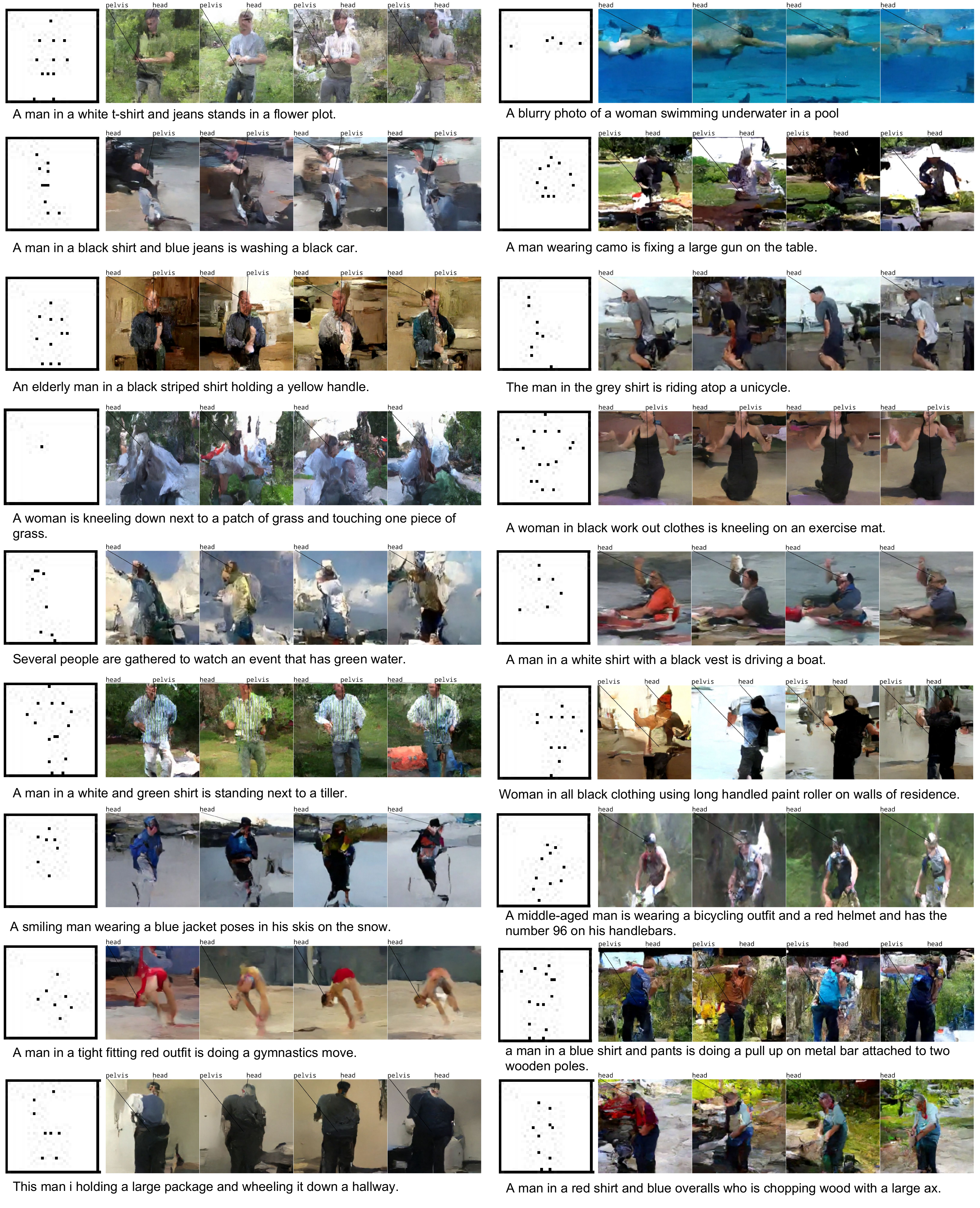}
\vspace{-0.15in}
\caption{Additional MPII samples randomly chosen from the validation set.\label{fig:mpii_appendix}}
\end{figure*}

\begin{figure*}[h]
\centering
\includegraphics[width=0.9\linewidth]{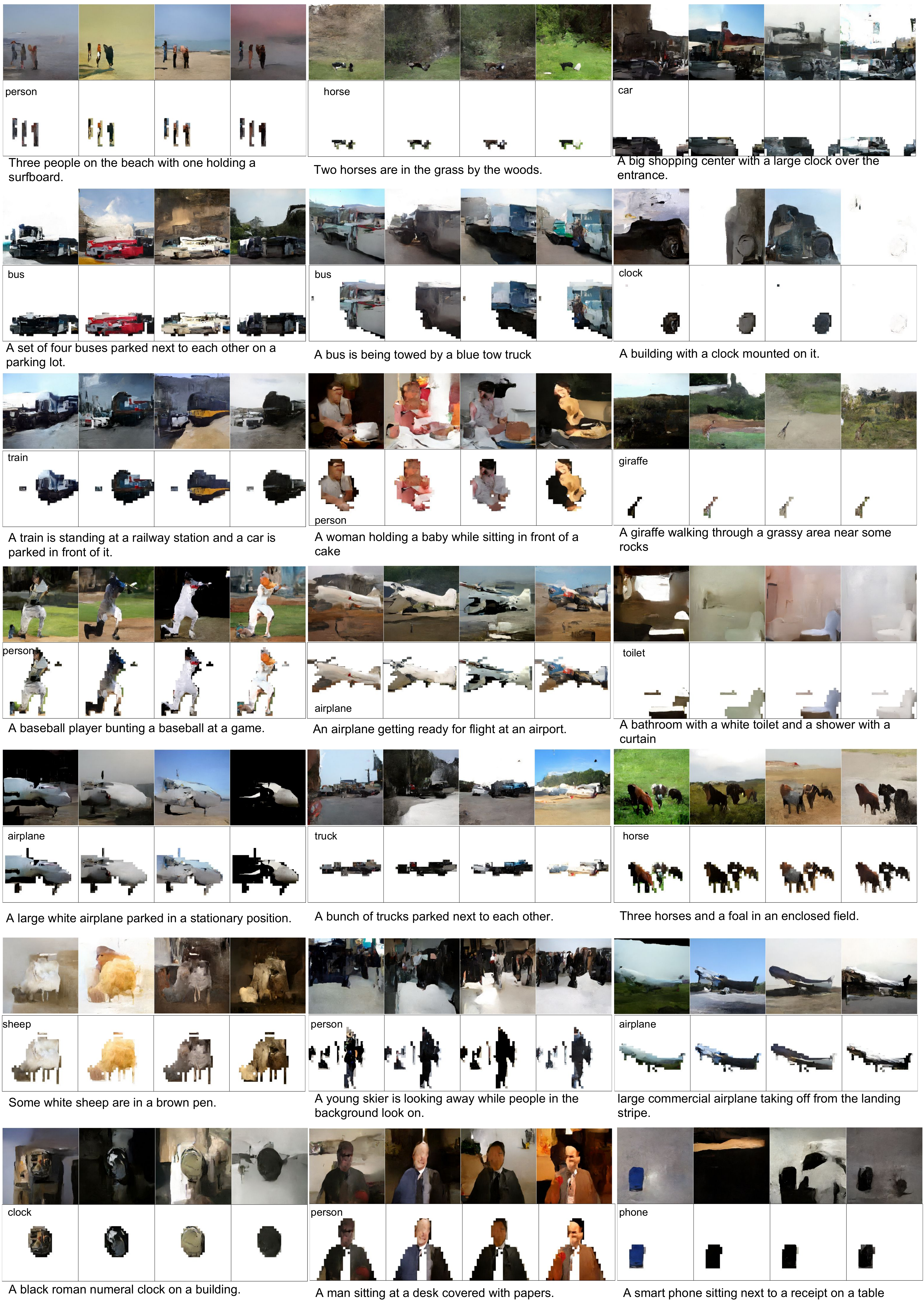}
\vspace{-0.15in}
\caption{Additional MS-COCO samples randomly chosen from the validation set.\label{fig:coco_appendix}}
\end{figure*}

\begin{figure*}[h]
\centering
\includegraphics[width=0.75\linewidth]{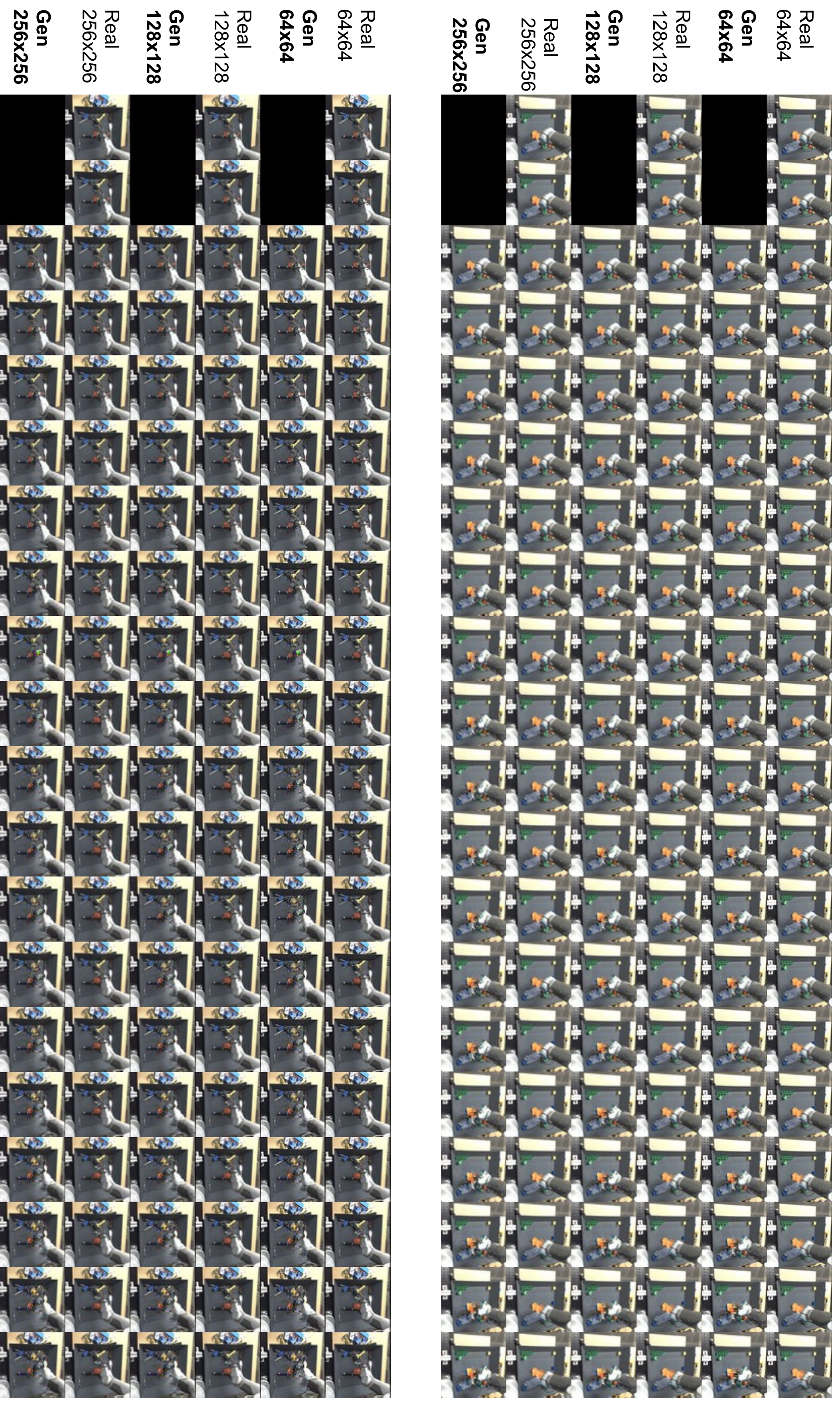}
\vspace{-0.15in}
\caption{Robot pushing videos at $64 \times 64$, $128 \times 128$ and $256 \times 256$.\label{fig:robot}}
\end{figure*}

\begin{figure*}[h]
\centering
\includegraphics[width=0.7\linewidth]{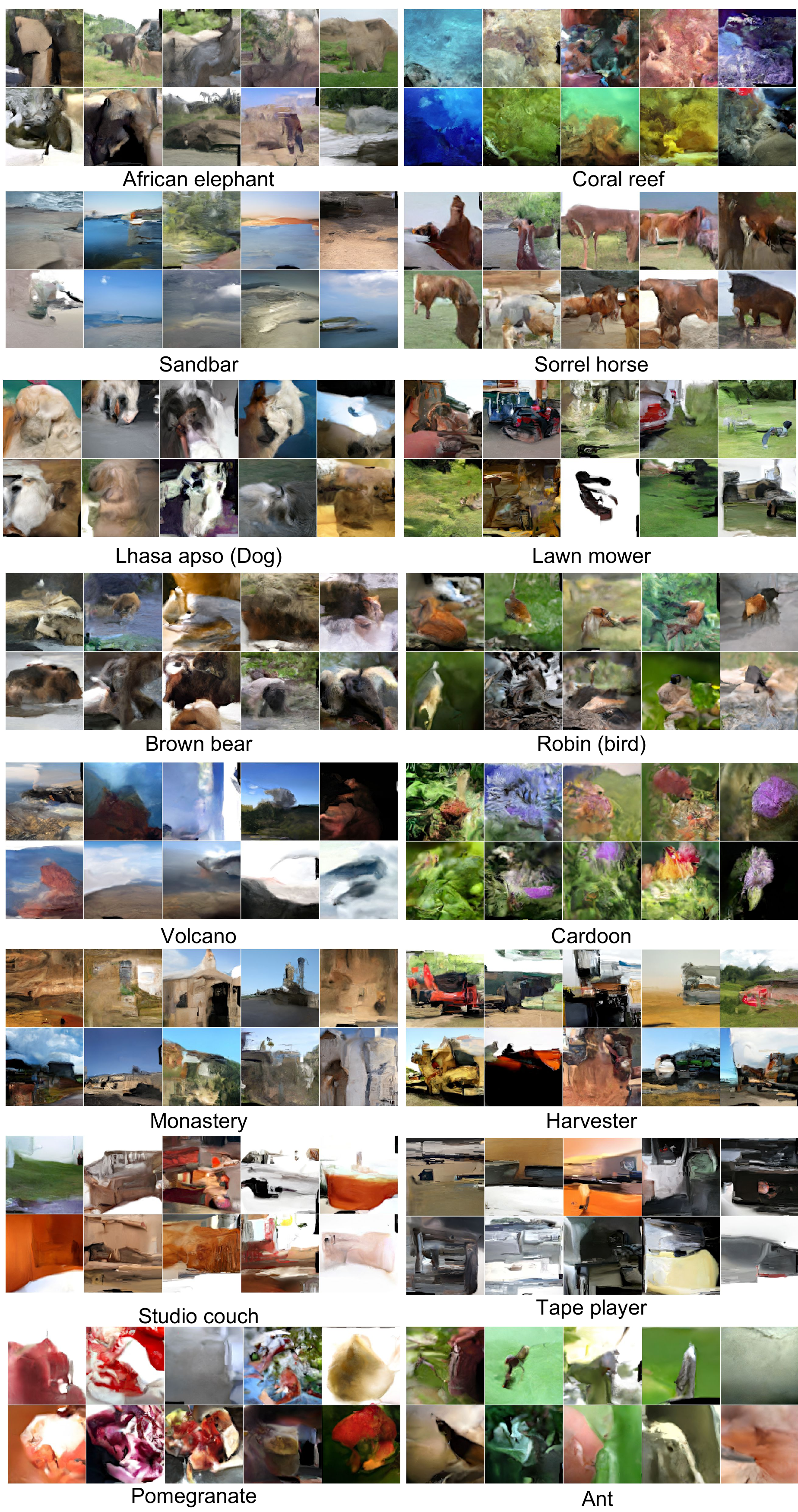}
\vspace{-0.15in}
\caption{Label-conditional $128 \times 128$ ImageNet samples.\label{fig:imagenet_appendix}}
\end{figure*}

\begin{figure*}[h]
\centering
\includegraphics[width=0.9\linewidth]{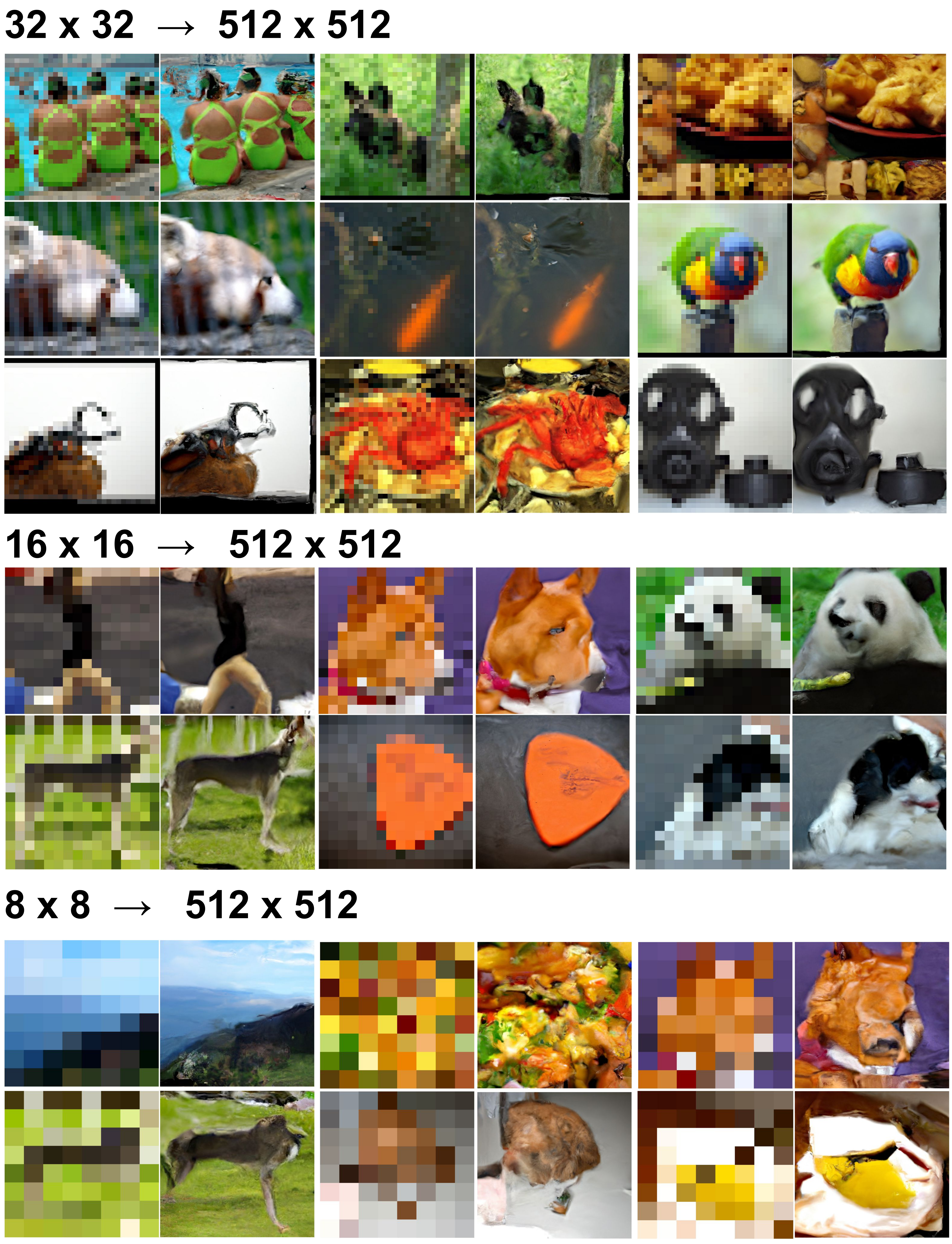}
\vspace{-0.15in}
\caption{Additional upscaling samples.\label{fig:imagenet_upscaling_appendix}}
\end{figure*}

\end{document}